%% file: main.tex
\documentclass[letterpaper]{article}

\usepackage[preprint]{neutralpreprint}
\usepackage[hyphens]{url}
\usepackage{graphicx}
\usepackage{natbib}
\usepackage{caption}
\usepackage{booktabs}
\usepackage{colortbl}
\usepackage{multirow}
\usepackage{amsmath}
\usepackage{amssymb}
\usepackage{microtype}

\newcommand{\method}{SPARK-SAM}
\newcommand{\samcore}{SAM2.1-Tiny}
\newcommand{\clip}{\operatorname{clip}_{[0,1]}}
\definecolor{SparkBestCyan}{HTML}{67C9D4}
\definecolor{SparkSecondOrange}{HTML}{F2AE72}
\newcommand{\SparkBest}[1]{\cellcolor{SparkBestCyan}\bfseries #1}
\newcommand{\SparkSecond}[1]{\cellcolor{SparkSecondOrange}\bfseries #1}

\title{SPARK-SAM: Learning How to Prompt and Respond for Infrared Small Target Segmentation}
\author{Aji Mao, Zhenming Peng, Bailin Mu, Tian Pu}
\affiliations{School of Information and Communication Engineering,\\
University of Electronic Science and Technology of China}

\begin{document}

\maketitle

\begin{abstract}
Promptable segmentation models provide a reusable interface, but direct transfer to automatic infrared small-target segmentation (IRSTD) exposes a mismatch between spatial prompts and target-domain mask responses.
In a diagnostic using target-covering loose-box prompts deterministically derived from test reference masks, the best official SAM2.1 results are only 4.69\%, 1.64\%, and 2.28\% IoU on NUAA-SIRST, NUDT-SIRST, and IRSTD-1K.
We introduce \method\ (\textbf{S}elf-\textbf{P}rompt \textbf{A}daptation with \textbf{R}esponse \textbf{K}nowledge for SAM), which learns target-domain response knowledge and conditions the decoder through an image-conditioned joint self-prompt state.
Training combines benchmark-mask supervision with reliability-aware response guidance.
\method\ achieves 75.78\%, 86.49\%, and 68.34\% IoU with 0.726M additional parameters, ranking first on two benchmarks among 14 retrained SAM variants and adaptations evaluated as automatic image-to-mask methods.
The staged IRSTD-1K diagnostic shows that response adaptation reaches most of the final IoU before the predicted points acquire reliable target grounding.
Prompt supervision aligns the predicted prompt candidates with target locations, and frozen-weight interventions measure output sensitivity to the joint self-prompt state.
Matched ablations show consistent accuracy gains from response guidance and high-resolution prompt refinement across all three datasets.
Code is available at \url{https://github.com/Sakauma/SPARK-SAM}.
\end{abstract}

\section{Introduction}

Infrared small-target segmentation (IRSTD) asks a model to isolate targets that may occupy only a few pixels under low contrast, sensor noise, and clutter.
Task-specific networks address these conditions through local--global context, shape priors, low-rank decomposition, and background modeling \citep{dai2021acm,li2023dnanet,zhang2022isnet,yuan2024sctransnet,wu2024rpcanet,wu2024l2sknet,liu2025forgetting}.
Promptable foundation models such as the Segment Anything Model (SAM) and SAM2 \citep{kirillov2023segment,ravi2025sam2} offer a different opportunity: a reusable encoder--prompt encoder--mask decoder interface could reduce dependence on a bespoke segmentation architecture.
Two properties of IRSTD make direct transfer especially demanding.
First, infrared small targets differ in appearance and scale from objects in natural images.
Second, IRSTD typically requires automatic image-to-mask prediction, while SAM2 relies on externally supplied spatial prompts.

\begin{figure}[t]
  \centering
  \includegraphics[width=\linewidth]{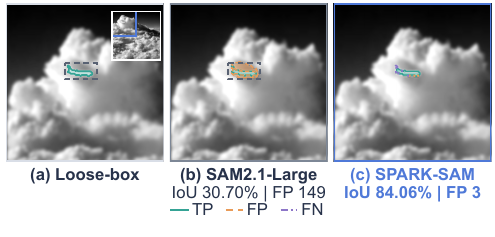}
  \caption{Prompt--response gap in direct SAM2.1 transfer. A loose-box prompt deterministically derived from the test reference mask covers the target, yet the unadapted response contains extensive false positives.}
  \label{fig:prompt-response-gap}
\end{figure}

The benchmarks provide pixel-level reference masks rather than separate bounding-box annotations; box-related training targets are derived from the corresponding training masks.
To test whether guaranteed target coverage is sufficient for direct transfer, we conduct a diagnostic with mask-derived bounding-box prompts: each test reference mask is deterministically converted into an external target-covering loose-box prompt for four frozen official SAM2.1 models.
The best dataset-level IoU is only 4.69\% on NUAA-SIRST, 1.64\% on NUDT-SIRST, and 2.28\% on IRSTD-1K.
High recall coexists with low precision and widespread false-positive activation.
Under this diagnostic protocol, target coverage alone is insufficient: the boxes cover all target pixels, yet the pretrained prompt-to-mask pathway produces similar responses for weak infrared signatures and clutter.
We call this failure the \emph{prompt--response gap}, which motivates adaptation of target-domain prompt-to-mask behavior.
Figure~\ref{fig:prompt-response-gap} contrasts the resulting unadapted response with \method's image-only prediction.

The gap motivates joint adaptation of SAM's conditional mask response and the prompt state that invokes it.
We use \emph{response knowledge} to denote the target-domain conditional prompt-to-mask behavior of the adapted SAM pathway.
For an image and prompt state, this behavior converts weak infrared evidence into compact target masks while suppressing clutter responses.
Benchmark annotations supervise the desired response, and reliability-filtered auxiliary responses are included as structural and ranking targets.
Self-prompt adaptation predicts an image-conditioned joint prompt state that invokes the learned response without external prompts.
We track mask accuracy and prompt grounding across training stages, then test decoder use through frozen-weight interventions.

We introduce \method\ (\textbf{S}elf-\textbf{P}rompt \textbf{A}daptation with \textbf{R}esponse \textbf{K}nowledge for SAM).
The framework retains SAM2's encoder--prompt encoder--decoder structure.
Response-knowledge adaptation learns the target-domain conditional behavior of the prompt-to-mask pathway.
An image-conditioned generator combines encoded coordinates, location-sampled tokens, residual prompt context, and a compact box into a joint self-prompt state; reliability-aware response guidance supplies training targets, and high-resolution refinement adds an extra candidate.

Across three IRSTD datasets, \method\ adds 0.726M parameters to \samcore\ and reaches 75.78\%, 86.49\%, and 68.34\% IoU.
Stagewise diagnostics track mask accuracy and prompt grounding, and frozen-weight interventions measure output dependence on the joint self-prompt state.
Matched ablations quantify the gains from the complete response-guidance pipeline and high-resolution prompt refinement.

Our contributions are threefold:
\begin{itemize}
  \item We identify a prompt--response gap in direct SAM2 transfer to IRSTD: even target-covering loose-box prompts derived from test reference masks yield poor masks, motivating adaptation of target-domain prompt-to-mask behavior.
  \item We formulate automatic IRSTD with SAM as the composition of a target-domain conditional response and an image-conditioned prompt state. \method\ learns this response as response knowledge and generates a target-grounded joint state for automatic image-to-mask inference.
  \item Stagewise diagnostics show that response adaptation reaches most of the final IoU before reliable point grounding emerges. Spatial measurements and frozen-weight interventions show that self-prompt learning aligns candidates with targets and forms a joint state used by the decoder. Matched ablations quantify the gains from response guidance and prompt refinement.
\end{itemize}

\section{Related Work}

\paragraph{Infrared small-target segmentation.}
IRSTD methods must preserve weak target signals while suppressing clutter.
Representative designs combine cross-level context and attention \citep{dai2021acm,li2023dnanet}, shape-aware supervision and transformer interactions \citep{zhang2022isnet,yuan2024sctransnet}, sparse-background decomposition \citep{wu2024rpcanet,wu2025rpcanetpp}, target-sensitive kernels \citep{wu2024l2sknet}, or explicit background and noise suppression \citep{liu2025forgetting,yuan2026seeing}.
These task-specific architectures establish strong accuracy, but effective transfer of a general promptable model to IRSTD remains underexplored.

\paragraph{Promptable segmentation beyond natural images.}
SAM \citep{kirillov2023segment} and SAM2 \citep{ravi2025sam2} separate visual encoding, prompt encoding, and mask decoding.
Extensions improve mask fidelity or efficiency \citep{ke2023hqsam,zhao2023fastsam,xiong2024efficientsam,xiong2024efficienttam,bonazzi2025picosam2}, while downstream variants introduce adapters, encoder-based prediction heads, or domain-specific specialization \citep{chen2023samadapter,chen2024sam2adapter,xiong2024sam2unet,xiong2025sam2unext,yao2025remotesam}.
IRSAM adapts SAM to IRSTD through Perona--Malik diffusion blocks and a granularity-aware decoder \citep{zhang2024irsam}.
SAMamba provides a closely matched SAM-family IRSTD adaptation through hierarchical state-space modeling \citep{xu2025samamba}.
\method\ couples response-knowledge adaptation with image-conditioned prompting through SAM2's prompt-conditioned decoder.

\paragraph{Automatic prompting and auxiliary knowledge.}
The dependence of SAM on user-provided points or boxes has motivated methods that infer spatial prompts directly from image features.
Prior work applies automatic prompting to nuclei, medical, and shadow segmentation \citep{sun2024segmentanynuclei,zhao2025selfprompt,jie2025shadowadapter}.
SAM-SPL learns prompts for single-frame infrared imagery \citep{fu2025unified}, while related work also studies sequential infrared segmentation \citep{dan2025oneshot}.
These studies establish learned prompt localization as a useful interface.
IRSTD additionally requires adaptation of the target-domain mask response.

Auxiliary knowledge provides a complementary means of adapting segmentation models.
Classical distillation transfers soft responses from a stronger model \citep{hinton2015distilling}, while structured segmentation distillation preserves spatial relations in dense prediction \citep{liu2019structuredkd}.
EdgeSAM further incorporates prompt information into the distillation process for efficient promptable segmentation \citep{zhou2025edgesam}.
Benchmark masks anchor \method's optimization.
Reliability-filtered frozen-model responses provide structural and ranking cues to the image-conditioned prompt generator and prompt-conditioned decoder.

\section{Method}

\subsection{Self-Prompting Closes the Transfer Loop}

Let $I\in\mathbb{R}^{H\times W}$ be an infrared image and $G\in\{0,1\}^{H\times W}$ its benchmark-provided pixel-level foreground mask.
During training, $G$ defines the desired response and foreground support; inference maps a new image directly to a binary mask through internal spatial prompts.
Box targets needed for training are derived from $G$, whereas normal inference predicts the complete prompt state from $I$ alone.
Denote the native SAM2 image encoder, prompt encoder, and mask decoder by $E_{\theta}$, $P_{\rho}$, and $D_{\omega}$.
From $Z=E_{\theta}(I)$, a self-prompt module $Q_{\phi}$ forms point coordinates, location-sampled tokens, residual prompt context, and a compact box; a refinement head $R_{\psi}$ can append a candidate from the high-resolution encoder map $Z_h$:
\begin{equation}
  \begin{aligned}
  \mathcal{P}_{I}&=Q_{\phi}\!\left(Z,R_{\psi}(Z_h)\right)
      =(\{\widehat u_j\},b,T_{\mathrm{res}},\{t_j\}),\\
  S_I&=\left[P_{\rho}(b,\{\widehat u_j\}),T_{\mathrm{res}},\{t_j\}\right],\\
  \widehat{M}&=D_{\omega}(Z,S_I).
  \end{aligned}
  \label{eq:forward}
\end{equation}
\begin{figure*}[t]
  \centering
  \includegraphics[width=\textwidth]{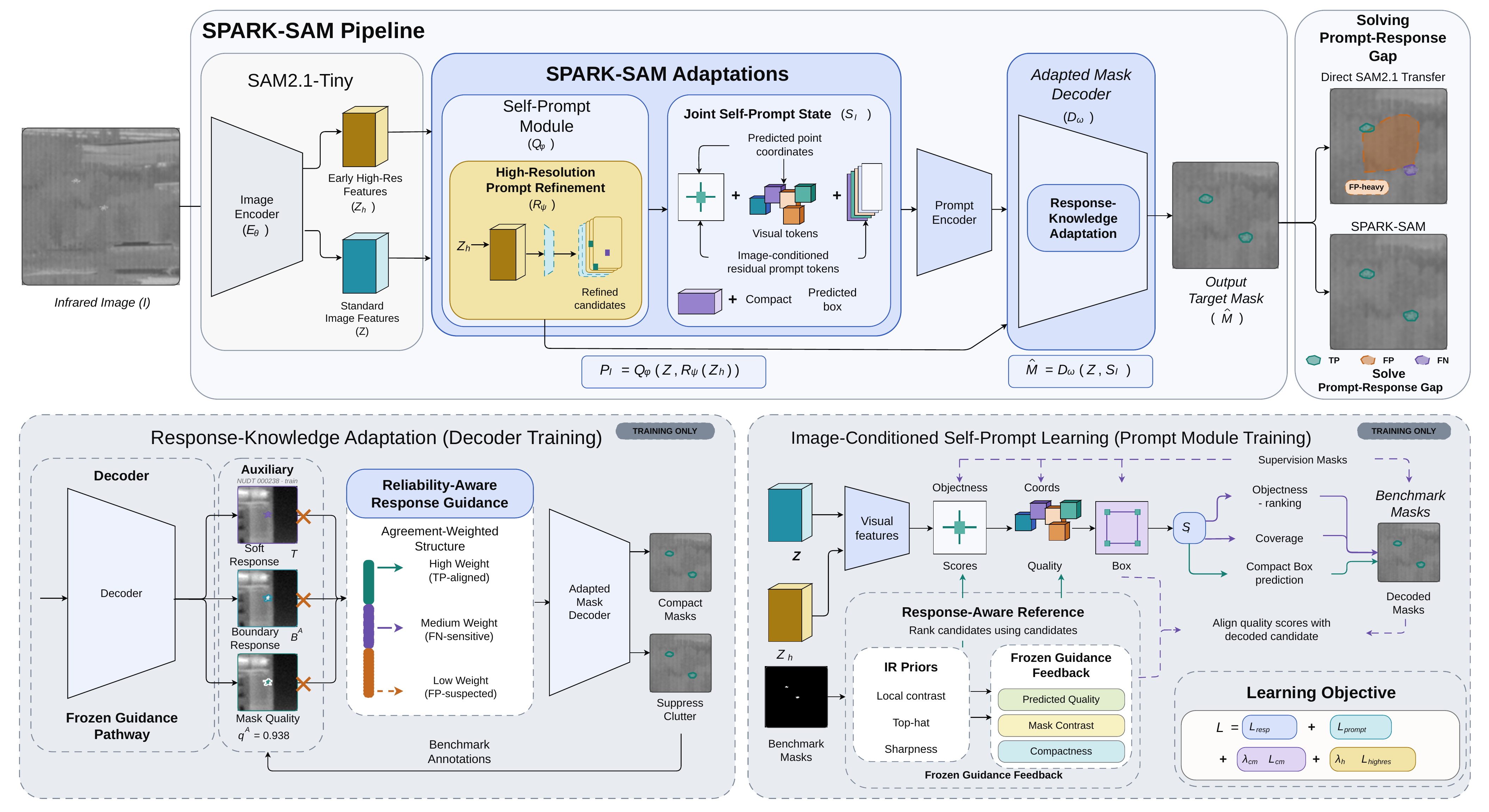}
  \caption{Overview of \method. Response-knowledge adaptation learns target-domain conditional prompt-to-mask behavior, and self-prompt adaptation conditions the decoder through a joint self-prompt state containing encoded coordinates, location-sampled tokens, residual prompt context, and a compact box. Reliability-aware response guidance supports training, and high-resolution encoder maps provide an additional prompt candidate.}
  \label{fig:framework}
\end{figure*}

Figure~\ref{fig:framework} summarizes this closed image-to-prompt-to-mask process, which we call \emph{self-prompting}: response knowledge determines the mask response, and the joint self-prompt state locates the evidence that invokes it.

\subsection{Reliability Weighting Filters Response Guidance}

The benchmark mask anchors target-domain response knowledge through binary cross-entropy (BCE) and Dice losses:
\begin{equation}
  \mathcal{L}_{\mathrm{seg}}
  =\operatorname{BCE}(\widehat{M},G)+\operatorname{Dice}(\widehat{M},G),
  \label{eq:seg}
\end{equation}
with a focal penalty $\mathcal{L}_{\mathrm{bg}}$ on foreground probability outside $G$.

The response-guidance pipeline $A$ returns a soft mask $T$, boundary distribution $B^A$, and mask-quality estimate $q^A$.
With $y_x^A=\mathbb{1}[T_x\geq\tau]$, annotations assign each pixel a reliability weight
\begin{equation}
 r_x=\begin{cases}
 \rho_a, & y_x^{A}=G_x,\\
 \rho_s, & G_x=0\wedge y_x^{A}=1,\\
 \rho_d, & \text{otherwise},
 \end{cases}
 \label{eq:reliability}
\end{equation}
where $\rho_a>\rho_d>\rho_s\geq0$ mark aligned, default, and suspected false-positive responses.
The weighted response loss is
\begin{equation}
  \begin{aligned}
  \mathcal{L}_{\mathrm{mask}}^{A}
    &=\frac{\sum_x r_x\ell_x^A}{\sum_x r_x},\\
  \ell_x^A&=\operatorname{BCE}(\widehat M_x,T_x)
      +(\sigma(\widehat M_x)-T_x)^2,
  \end{aligned}
  \label{eq:response-mask}
\end{equation}
Adding shape loss $\mathcal{L}_{\mathrm{bnd}}^{A}$ and calibration loss $\mathcal{L}_{\mathrm{qual}}^{A}$ gives
\begin{equation}
  \begin{aligned}
  \mathcal{L}_{\mathrm{resp}}
  ={}&\lambda_{g}\mathcal{L}_{\mathrm{seg}}
   +\lambda_{b}\mathcal{L}_{\mathrm{bg}}
   +\lambda_{m}\mathcal{L}_{\mathrm{mask}}^{A}
   \\
   &+\lambda_{e}\mathcal{L}_{\mathrm{bnd}}^{A}
   +\lambda_{q}\mathcal{L}_{\mathrm{qual}}^{A}.
  \end{aligned}
  \label{eq:response}
\end{equation}
This objective anchors the response in $G$ and filters unreliable guidance.
\subsection{The Joint State Invokes Response Knowledge}

The self-prompt module $Q_\phi$ maps $Z$ to an objectness logit map $o$, offsets $\Delta$, candidate scores $\{c_j\}$, candidate-mask quality estimates, and a compact box $b$.
It selects candidate coordinates and samples their local evidence:
\begin{equation}
  \begin{aligned}
  \widehat{\mathcal C}
    &=\{\widehat u_j=x_j+\Delta_j\mid
        x_j\in\operatorname{TopK}_{K}(o)\},\\
  t_j&=W_{\mathrm{loc}}\!\left(\mathcal{B}(Z,\widehat u_j)\right),
  \end{aligned}
  \label{eq:self-prompt-candidates}
\end{equation}
where $\mathcal{B}$ bilinearly samples $Z$ and $W_{\mathrm{loc}}$ is a LayerNorm--multilayer perceptron (MLP) projector.
The residual sparse tokens $T_{\mathrm{res}}$ share image-conditioned context across candidates.
Brackets denote concatenation in Equation~\eqref{eq:forward}; the prompt encoding of coordinates and box, $T_{\mathrm{res}}$, and the local tokens $\{t_j\}$ form $S_I$.
The coordinate/token pairs ground local evidence, and the box supplies its extent.

Training combines benchmark-mask supervision with a response-guidance reference.
The estimator $A_p$ yields an initial box and $K_A$ non-maximum-suppressed candidates $\{u_j\}$, then collects each candidate's infrared cues in $\mathbf z_j^p=[a_j,L_j,U_j,\pi_j,\varphi_j]^\top$: objectness, local contrast, top-hat response, peak sharpness, and high-frequency evidence.
A fixed vector assigns the prior score $p_j=\mathbf w_p^\top\mathbf z_j^p$.
For the $K_F$ largest $p_j$, the frozen guidance model decodes a positive point into mask $M_j$ and quality $q_j$, then collects feedback $\mathbf z_j^f=[\clip(q_j),\chi_j,\eta(v_j),\kappa_j,\zeta_j]^\top$ from mask contrast, area compatibility, compactness, and point--mask center agreement.
Here, $v_j=|\mathbb 1[M_j>0.5]|/(HW)$ is the normalized mask area.
A second fixed vector assigns $f_j=\mathbf w_f^\top\mathbf z_j^f$.
Let $\mathcal J_F$ index the decoded candidates. With $\mu\in[0,1]$, the reference selection is
\begin{equation}
  \begin{aligned}
  s_j&=\begin{cases}
    \mu p_j+(1-\mu)f_j, & j\in\mathcal{J}_F,\\
    p_j, & j\notin\mathcal{J}_F,
  \end{cases}\\
  j^{\star}&=\arg\max_j s_j,
  \qquad u^{\star}=u_{j^{\star}}.
  \end{aligned}
  \label{eq:reference-selection}
\end{equation}
The point $u^\star$ recenters $A_p$'s box.
For each scale $\gamma\in\Gamma$, let $b_\gamma$ denote the scaled box, $\mathcal N_\gamma$ its surrounding-background points, and $\beta_\gamma$ the response score from decoding $(u^\star,b_\gamma,\mathcal N_\gamma)$.
Let $\bar f_{j^\star}$ equal $f_{j^\star}$ for $j^\star\in\mathcal J_F$ and zero otherwise.
With margin $\delta\geq0$, calibration selects
\begin{equation}
  \begin{aligned}
  \gamma^\star&=\arg\max_{\gamma\in\Gamma}\beta_\gamma,\quad
  g=\mathbb 1[\beta_{\gamma^\star}\geq\bar f_{j^\star}+\delta],\\
  (\mathcal P^A,b^\star)&=
  \begin{cases}
  ((u^\star,b_{\gamma^\star},\mathcal N_{\gamma^\star}),b_{\gamma^\star}),
      & g=1,\\
  ((u^\star),b_G),&g=0,
  \end{cases}
  \end{aligned}
  \label{eq:box-calibration}
\end{equation}
where $b_G$ is the training-mask-derived fallback target.
Decoding $\mathcal P^A$ yields logits $L^A$ and quality $q^A$; the response, boundary, and spatial targets are
\begin{equation}
  \begin{aligned}
  T&=\sigma(L^A),\notag\\
  B^A&=\clip\!\left(\operatorname{MaxPool}_{r_b}(T)
     -\operatorname{MinPool}_{r_b}(T)\right),\\
  O^\star(x)&=\exp\!\left(-\frac{\lVert x-u^\star\rVert_2^2}{2\sigma_o^2}\right).
  \end{aligned}
  \label{eq:response-target-construction}
\end{equation}
Here, $r_b$ is the local pooling window and $\sigma_o$ is the Gaussian width.
The prompt losses combine response-guidance targets with benchmark grounding: point and box terms use $(u^\star,b^\star)$, while ranking and coverage place candidates on $G$.
Candidate-mask quality regresses to soft IoU with $G$, and $A_p$ supplies a dense prior $\widetilde O=A_p(I)$ from objectness, box, and ranking targets derived from the training masks.
\begin{equation}
  \begin{aligned}
  \mathcal{L}_{\mathrm{prompt}}
  ={}&\lambda_o\mathcal{L}_{\mathrm{obj}}(o,O^{\star})
  +\lambda_p\mathcal{L}_{\mathrm{point}}\\
  &+\lambda_{box}\mathcal{L}_{\mathrm{box}}(b,b^{\star})
  +\lambda_r\mathcal{L}_{\mathrm{rank}}(\{c_j\},G)\\
  &
  +\lambda_c\mathcal{L}_{\mathrm{coverage}}
  +\lambda_d\mathcal{L}_{\mathrm{dense}}(o,\widetilde{O}).
  \end{aligned}
  \label{eq:prompt}
\end{equation}

\subsection{High-Resolution Residuals Refine Localization}

Low-resolution objectness prediction reduces computation but coarsens localization for targets that span only a few pixels.
Writing the learned low-resolution map as $o_\ell=o$, a zero-initialized head $R_\psi$ refines it from an early high-resolution encoder map $Z_h$ with residual scale $\alpha$:
\begin{equation}
  o_h=\operatorname{Up}(o_{\ell})+\alpha R_{\psi}(Z_h).
  \label{eq:highres}
\end{equation}
The prompt set retains $\mathcal C_\ell=\operatorname{TopK}_{K_\ell}(o_\ell)$ and the $K_r$ highest-ranked elements of $\mathcal C_h=\operatorname{TopK}_{K_h}(o_h)$.
Zero initialization starts from $\operatorname{Up}(o_\ell)$; retaining $\mathcal C_\ell$ preserves coarse candidates while $\mathcal C_h$ adds only $K_r$ fine hypotheses.
Objectness, ranking, and coverage losses form $\mathcal L_{\mathrm{highres}}$, and $\mathcal L_{\mathrm{cm}}$ denotes the candidate-mask quality regression.
\begin{equation}
  \mathcal L=\mathcal L_{\mathrm{resp}}+\mathcal L_{\mathrm{prompt}}
  +\lambda_{cm}\mathcal L_{\mathrm{cm}}
  +\lambda_h\mathcal L_{\mathrm{highres}}.
  \label{eq:total}
\end{equation}

\begin{figure}[!b]
  \centering
  \includegraphics[width=\linewidth]{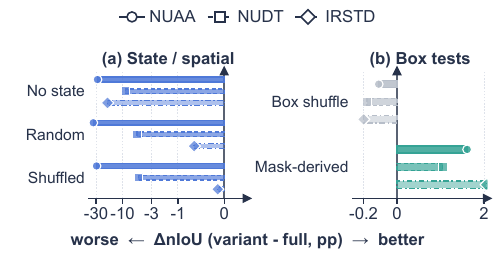}
  \caption{Frozen-weight dependence on the joint self-prompt state. Bars report variant-minus-full-model $\Delta\mathrm{nIoU}$ in percentage points; panel-specific symmetric-log horizontal scales expand the near-zero box effects. Complete removal deletes $S_I$; spatial interventions preserve $T_{\mathrm{res}}$ and the box, while box interventions preserve $T_{\mathrm{res}}$ and the point/local-token branch. Post-adaptation mask-derived box replacement measures residual box-conditioned headroom.}
  \label{fig:prompt-dependence}
\end{figure}

\section{Experiments}

\subsection{Datasets, Metrics, and Protocol}

We evaluate on NUAA-SIRST \citep{dai2021acm}, NUDT-SIRST \citep{li2023dnanet}, and IRSTD-1K \citep{zhang2022isnet}.
Each benchmark provides pixel-level masks without separate box annotations.
The train/validation/test splits contain 256/85/85 images for NUAA-SIRST, 797/265/265 for NUDT-SIRST, and 601/200/200 for IRSTD-1K.
Training masks supervise mask prediction and the derived objectness, box, and ranking targets without additional box labels.
Validation masks select checkpoints, mask thresholds, and multimask outputs; test masks support scoring.
We train each retrained method per dataset.
At inference, automatic methods receive infrared images alone and run their image-to-mask pathways at native resolution.
All methods use the same validation-only selection rule and shared evaluator.

The four frozen official SAM2.1 models form a separate direct-transfer diagnostic.
For each test image, we derive an external loose box from the reference-mask union box, pad each side by $\max(0.15s,2)$ pixels for tight-box extent $s$, cap the expanded extent at $2.0s$, and clip it to the image.
The box covers every target pixel, and each frozen model predicts only the mask.
The evaluator computes this box from the reference annotation; no user or model predicts it.
This privileged diagnostic tests whether guaranteed target coverage closes the prompt--response gap.

Dataset-level IoU equals summed test-set intersections divided by summed unions; normalized IoU (nIoU) averages image-level IoUs \citep{dai2021acm}.
We also report global F1, precision, recall, and false-alarm pixels per megapixel (FApx/MP).
The comparisons cover eight specialized IRSTD networks, four official SAM2.1 scales, thirteen retrained SAM variants and adaptations, and \method.
The staged IRSTD-1K diagnostic tracks mask accuracy and prompt grounding, while matched ablations measure response guidance and prompt refinement.

\subsection{Implementation Details}

We initialize $E_{\theta}$, $P_{\rho}$, and $D_{\omega}$ from the official \samcore\ weights and process min--max-normalized three-channel frames at $1024^2$ with the official transforms.
The frozen response-guidance pipeline pairs SAM2.1-Large with $A_p$, which we train on the training split and select on the validation split.
$A_p$ learns objectness, box, and ranking targets derived from training masks.
It produces $T$, $B^{A}$, $q^{A}$, $(u^{\star},b^{\star})$, and dense objectness supervision for $Q_{\phi}$.
We use $\tau=0.5$ and set the reliability coefficients $(\rho_a,\rho_d,\rho_s)$ to $(1.0,0.75,0.25)$.
Both candidate pools retain four locations ($K_{\ell}=K_h=4$), and the decoder adds the highest-ranked refined point ($K_r=1$) to the four low-resolution points.
Joint adaptation uses AdamW with learning rate $10^{-5}$, weight decay $0.01$, effective batch size 24, and a $0.1$ image-encoder learning-rate multiplier.
Calibration updates the mask decoder and local-token projector with the encoders and prompt head fixed.
We then train high-resolution refinement for 10 epochs at a smaller learning rate while fixing the preceding modules.
Validation performance selects the final checkpoint and mask threshold.

\begin{figure}[!b]
  \centering
  \includegraphics[width=\linewidth]{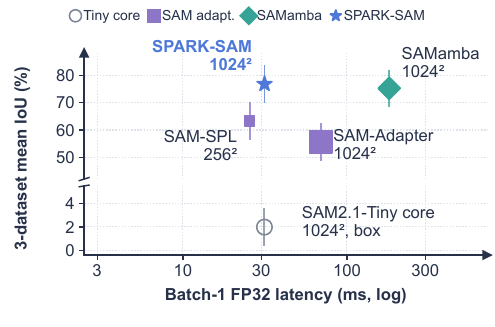}
  \caption{Mean IoU across NUAA-SIRST, NUDT-SIRST, and IRSTD-1K versus audited batch-1 FP32 latency. The broken vertical axis separates the low-IoU Tiny-core anchor from the adapted methods. Marker shapes distinguish method families, marker sizes denote model complexity, and labels indicate native input resolution.}
  \label{fig:efficiency}
\end{figure}

\subsection{Comparison with SAM2 and Related Adaptations}

\input{tables/sam_comparison}

Table~\ref{tab:sam-comparison} quantifies the prompt--response gap under the direct-transfer diagnostic with mask-derived box prompts.
All four official SAM2.1 scales remain below 5\% IoU despite target-covering boxes.
The gap therefore persists from Tiny through Large and motivates target-domain response adaptation.
Among fourteen retrained SAM variants and adaptations, \method\ reaches 75.78\%, 86.49\%, and 68.34\% IoU, ranking second on NUAA-SIRST and first on NUDT-SIRST and IRSTD-1K.
SAMamba leads on NUAA-SIRST.
Pretrained scale leaves the target-domain response failure unresolved.

\subsection{Comparison with Specialized IRSTD Networks}

\input{tables/specialized_comparison}

Table~\ref{tab:specialized-comparison} compares \method\ with eight task-specific networks under the same split and evaluation protocol.
\method\ ranks fourth on NUAA-SIRST, fifth on NUDT-SIRST, and second on IRSTD-1K.
SCTransNet retains the highest specialized-model IoU on all three datasets.
The comparison supports effective SAM2 transfer while preserving the advantage of task-specific models.

\subsection{Response Acquisition and Self-Prompt Invocation}

\input{tables/mechanism_analysis}

\paragraph{Response adaptation precedes reliable prompt grounding.}
For image $i$, let $\mathcal{C}_i^K$ denote the final candidate set and $\overline{u}$ a candidate mapped to the original resolution, rounded, and clipped to the image domain.
We define
\begin{equation}
  \operatorname{Hit@K}
  =\frac{1}{N}\sum_{i=1}^{N}
   \mathbb{1}\!\left[\exists\,u\in\mathcal{C}_i^K:G_i(\overline{u})=1\right].
  \label{eq:hit-at-k}
\end{equation}
Hit@$K$ measures the fraction of images with at least one candidate on the reference foreground.
The diagnostic uses four low-resolution candidates before refinement ($K=4$) and adds the highest-ranked refined candidate afterward ($K=5$).
Response adaptation reaches 68.01\% IoU at Hit@$K=0.005$.
Prompt supervision raises Hit@$K$ to 0.540, adds 2.89 precision points, and reduces FA by 10.11 px/MP with a $0.35$-point IoU decrease.
Refinement reaches 68.34\% IoU and Hit@$K=0.955$.
In this diagnostic, response learning supplies most mask accuracy, while self-prompt losses align candidates with target evidence and reduce false alarms.
Changes across stages separate response acquisition from prompt grounding: mask accuracy emerges first, followed by spatial alignment and false-alarm control.

\begin{figure}[!t]
  \centering
  \includegraphics[width=0.92\linewidth]{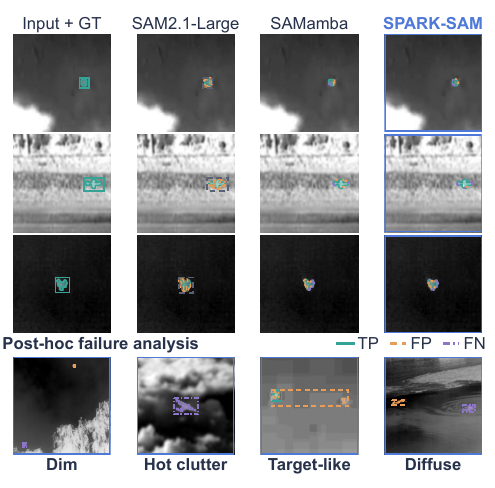}
  \caption{Qualitative comparison and failures. Columns: input+ground truth (GT), SAM2.1-Large, SAMamba, and SPARK-SAM. Colors: true/false positives and false negatives (TP/FP/FN). Bottom: dim targets, hot clutter, target-like distractors, and diffuse responses.}
  \label{fig:qualitative}
\end{figure}

\paragraph{Decoded masks depend on the joint self-prompt state.}
We keep all model weights fixed and alter the joint self-prompt state $S_I$.
Removing $S_I$ deletes the encoded coordinates and box, $T_{\mathrm{res}}$, and local tokens.
Spatial interventions retain $T_{\mathrm{res}}$ and the box while changing coordinates and resampling local tokens; box interventions retain $T_{\mathrm{res}}$ and the point/local-token branch.
The mask-derived replacement retains learned points and local tokens but substitutes the test-reference union box to measure residual box-conditioned headroom.
For each moved point, we resample its local token at the new location.
These replacements support diagnosis only; normal inference uses the predicted joint state.
Resampling preserves the coupling between each coordinate and its local appearance token.

Table~\ref{tab:mechanism-analysis}(a) and Figure~\ref{fig:prompt-dependence} show frozen-weight dependence on the joint prompt state.
Complete removal lowers mean nIoU by 8.84--28.27 points across the three datasets.
Randomizing or shuffling the spatial branch lowers nIoU on every dataset by 0.65--33.63 and 0.14--29.41 points, respectively; box shuffling lowers nIoU by 0.11--0.20 points.
Mask-derived box replacement raises IoU by 1.57, 2.48, and 1.77 points on NUAA-SIRST, NUDT-SIRST, and IRSTD-1K.
The coordinate-conditioned local-token channel supplies spatial information beyond coarse box context, while the replacement test exposes remaining localization headroom.
The effect size varies across datasets, but every spatial intervention degrades nIoU.

\paragraph{Guidance and refinement improve the final model.}
Removing the response-guidance pipeline lowers IoU by 12.35--23.87 points across the three datasets.
Removing refinement lowers IoU by 0.50--1.52 points and reduces Hit@$K$ from 0.955 to 0.540.
Refinement also changes the candidate count from four to five, so Hit@$K$ combines high-resolution localization with the additional candidate.
These matched ablations attribute the larger accuracy gain to the complete guidance pipeline and the smaller localization gain to refinement.

\subsection{Efficiency Analysis}

\method\ contains 39.688M parameters and adds 0.726M parameters, 6.80~MiB peak memory, and 6.134 profiled GFLOPs to \samcore.
The added modules raise mean latency from 31.400 to 31.451~ms, a 0.051~ms increase.
Across 600 measurements, mean/median/P95 latency reaches 31.451/30.716/37.573~ms, or 31.8 frames per second.

We benchmark batch-1 FP32 inference on an RTX~4090 after 20 warm-up iterations and three passes over the 200-image IRSTD-1K test set.
The measured path runs from an in-memory RGB array to a CPU binary mask and excludes disk I/O; the profiler reports recognized operations.
Table~\ref{tab:sam-efficiency} retains each method's native resolution.
At $1024^2$, \method\ processes 16 times the pixels of $256^2$ SAM-SPL, raises mean latency by 23.4\%, and improves IoU by 16.31, 19.34, and 5.09 points.
At the same $1024^2$ resolution, \method\ runs nearly six times faster than SAMamba.
Figure~\ref{fig:efficiency} therefore reports a native-resolution accuracy--latency comparison.
These measurements characterize each implementation's native inference path and retain resolution differences across methods.

\subsection{Qualitative Results and Failure Modes}

Figure~\ref{fig:qualitative} shows that target-covering boxes still produce broad clutter responses from frozen SAM2.1-Large.
Image-only \method\ produces compact masks around weak targets after adaptation and preserves several dim or irregular targets.
Failures remain on hot clutter, target-like distractors, and diffuse boundaries.
All methods use identical crops and TP/FP/FN colors.

\section{Conclusion}

\method\ addresses the prompt--response gap exposed by frozen SAM2.1 responses to test-mask-derived target-covering loose-box prompts.
It combines target-domain response knowledge with an image-conditioned joint self-prompt state for automatic inference.
Across three benchmarks, its automatic masks outperform most retrained SAM variants and adaptations (Table~\ref{tab:sam-comparison}).
The staged IRSTD-1K evaluation shows that response adaptation reaches most of the final IoU before reliable point grounding; frozen-weight interventions confirm that the decoder uses the joint self-prompt state (Figure~\ref{fig:prompt-dependence}).
We retrain methods separately per dataset and evaluate them on held-out test splits; post-hoc failure cases involve dim targets, hot clutter, target-like distractors, and diffuse responses (Figure~\ref{fig:qualitative}).
Future work will test \method\ in cross-dataset and temporal settings.

\bibliography{references}

\end{document}

%% file: tables/sam_comparison.tex
\begin{table*}[t]
  \centering
  \begingroup
  \scriptsize
  \setlength{\tabcolsep}{1.25pt}
  \renewcommand{\arraystretch}{0.88}
  \resizebox{\textwidth}{!}{%
  \begin{tabular}{@{}l *{3}{rrrrrr}@{}}
    \toprule
    & \multicolumn{6}{c}{NUAA-SIRST} & \multicolumn{6}{c}{NUDT-SIRST} & \multicolumn{6}{c}{IRSTD-1K} \\
    \cmidrule(lr){2-7}\cmidrule(lr){8-13}\cmidrule(lr){14-19}
    Method & IoU$\uparrow$ & nIoU$\uparrow$ & F1$\uparrow$ & Pr$\uparrow$ & Re$\uparrow$ & FA$\downarrow$
      & IoU$\uparrow$ & nIoU$\uparrow$ & F1$\uparrow$ & Pr$\uparrow$ & Re$\uparrow$ & FA$\downarrow$
      & IoU$\uparrow$ & nIoU$\uparrow$ & F1$\uparrow$ & Pr$\uparrow$ & Re$\uparrow$ & FA$\downarrow$ \\
    \midrule
    \multicolumn{19}{l}{\itshape Official SAM2.1 models with mask-derived box prompts} \\
    SAM2.1 Tiny (mask-derived box)~\citep{ravi2025sam2} & 2.99 & 37.24 & 5.80 & 3.00 & 90.08 & 18991.00 & 1.40 & 32.82 & 2.76 & 1.40 & 76.37 & 35925.50 & 1.57 & 32.55 & 3.09 & 1.58 & 76.55 & 11996.90 \\
    SAM2.1 Small (mask-derived box)~\citep{ravi2025sam2} & 4.69 & 37.58 & 8.96 & 4.71 & 92.47 & 13898.30 & 1.41 & 31.48 & 2.78 & 1.42 & 79.65 & 37098.30 & 1.70 & 31.57 & 3.35 & 1.71 & 80.55 & 11646.80 \\
    SAM2.1 Base+ (mask-derived box)~\citep{ravi2025sam2} & 4.00 & 38.09 & 7.68 & 4.01 & \SparkSecond{93.73} & 15652.80 & 1.64 & 33.22 & 3.23 & 1.64 & 86.97 & 34824.00 & 1.85 & 31.43 & 3.64 & 1.86 & 90.24 & 11980.10 \\
    SAM2.1 Large (mask-derived box)~\citep{ravi2025sam2} & 2.93 & 38.59 & 5.69 & 2.94 & 91.85 & 20560.50 & 1.16 & 33.76 & 2.28 & 1.16 & 80.16 & 45778.70 & 2.28 & 33.18 & 4.47 & 2.29 & 87.63 & 9388.80 \\
    \midrule
    \multicolumn{19}{l}{\itshape Retrained SAM variants and adaptations} \\
    EfficientSAM ViT-T~\citep{xiong2024efficientsam} & 2.79 & 45.31 & 5.43 & 2.80 & 88.65 & 19410.10 & 1.57 & 38.03 & 3.10 & 1.58 & 76.72 & 31985.90 & 1.73 & 39.95 & 3.40 & 1.73 & 83.73 & 11924.20 \\
    FastSAM-s~\citep{zhao2023fastsam} & 19.04 & 41.64 & 31.99 & 21.53 & 62.26 & 1380.00 & 10.12 & 42.64 & 18.38 & 10.68 & 65.98 & 3694.90 & 2.31 & 31.45 & 4.52 & 2.34 & 62.60 & 6558.80 \\
    HQ-SAM ViT-B~\citep{ke2023hqsam} & 8.71 & 43.14 & 16.03 & 8.78 & 91.91 & 7003.70 & 1.31 & 38.92 & 2.58 & 1.31 & \SparkBest{95.72} & 48261.90 & 1.86 & 42.98 & 3.65 & 1.86 & \SparkSecond{93.84} & 12427.20 \\
    SAM ViT-B~\citep{kirillov2023segment} & 1.74 & 41.53 & 3.42 & 1.74 & \SparkBest{95.87} & 35308.70 & 0.75 & 34.99 & 1.48 & 0.75 & 85.93 & 76518.90 & 0.81 & 34.31 & 1.60 & 0.81 & 77.71 & 23924.70 \\
    SAM-Adapter ViT-B~\citep{chen2023samadapter} & 66.35 & 70.44 & 79.77 & 76.74 & 83.06 & 135.40 & 57.44 & 58.99 & 72.97 & 63.36 & 86.03 & 333.00 & 43.23 & 46.01 & 60.36 & 46.71 & 85.28 & 244.40 \\
    SAM-SPL~\citep{fu2025unified} & 59.47 & 60.29 & 74.59 & 64.53 & 88.37 & 295.14 & 67.15 & 68.47 & 80.35 & \SparkSecond{93.48} & 70.45 & \SparkBest{32.88} & 63.25 & 55.36 & 77.49 & 79.51 & 75.57 & 48.92 \\
    EfficientTAM~\citep{xiong2024efficienttam} & 0.05 & 8.30 & 0.10 & 0.05 & 33.44 & 439515.00 & 0.06 & 3.58 & 0.11 & 0.06 & 25.20 & 304746.50 & 0.02 & 0.74 & 0.04 & 0.02 & 19.21 & 222976.60 \\
    PicoSAM2~\citep{bonazzi2025picosam2} & 0.06 & 0.06 & 0.12 & 0.06 & 91.77 & 914428.10 & 0.07 & 0.49 & 0.14 & 0.07 & 80.99 & 754228.70 & 0.03 & 0.03 & 0.06 & 0.03 & \SparkBest{95.36} & 815243.40 \\
    RemoteSAM~\citep{yao2025remotesam} & 0.25 & 19.64 & 0.50 & 0.25 & 81.76 & 180353.10 & 0.13 & 3.12 & 0.25 & 0.13 & 67.55 & 357846.90 & 0.25 & 13.47 & 0.50 & 0.25 & 71.82 & 72087.40 \\
    SAM2-Adapter~\citep{chen2024sam2adapter} & 0.02 & 1.16 & 0.04 & 0.02 & 6.18 & 179365.30 & 0.05 & 0.30 & 0.11 & 0.05 & 16.70 & 205474.30 & 0.02 & 0.79 & 0.03 & 0.02 & 16.21 & 247787.70 \\
    SAM2-UNeXt~\citep{xiong2025sam2unext} & 1.93 & 14.55 & 3.79 & 2.02 & 30.26 & 7559.00 & 1.30 & 7.61 & 2.57 & 1.37 & 20.93 & 10075.30 & 0.33 & 8.14 & 0.66 & 0.33 & 34.16 & 25538.20 \\
    SAM2-UNet~\citep{xiong2024sam2unet} & 5.24 & 25.42 & 9.96 & 5.41 & 62.43 & 6471.00 & 1.61 & 15.08 & 3.16 & 1.64 & 48.70 & 19605.50 & 0.62 & 10.16 & 1.23 & 0.62 & 45.45 & 18240.80 \\
    SAMamba~\citep{xu2025samamba} & \SparkBest{77.45} & \SparkBest{79.35} & \SparkBest{87.29} & \SparkBest{88.53} & 86.09 & \SparkBest{67.77} & \SparkSecond{81.47} & \SparkSecond{82.99} & \SparkSecond{89.79} & 86.09 & \SparkSecond{93.81} & 101.46 & \SparkSecond{67.01} & \SparkBest{65.18} & \SparkSecond{80.25} & \SparkSecond{81.39} & 79.14 & \SparkBest{45.47} \\
    \midrule
    \multicolumn{19}{l}{\itshape Ours} \\
    \textbf{SPARK-SAM} & \SparkSecond{75.78} & \SparkSecond{76.93} & \SparkSecond{86.22} & \SparkSecond{84.18} & \textbf{88.37} & \SparkSecond{100.88} & \SparkBest{86.49} & \SparkBest{87.75} & \SparkBest{92.75} & \SparkBest{93.63} & \textbf{91.90} & \SparkSecond{41.86} & \SparkBest{68.34} & \SparkSecond{64.47} & \SparkBest{81.19} & \SparkBest{81.53} & \textbf{80.86} & \SparkSecond{46.02} \\
    \bottomrule
  \end{tabular}%
  }
  \endgroup
  \caption{Segmentation comparison with official and retrained SAM variants. IoU, nIoU, F1, precision (Pr), and recall (Re) are percentages; FA is in px/MP. Official rows form a diagnostic with mask-derived box prompts: each frozen SAM2.1 model receives a target-covering loose box deterministically computed from the corresponding test reference mask and predicts only the mask. Retrained rows receive only infrared images at inference and use automatic image-to-mask pathways. Cyan and orange denote the best and second-best results.}
  \label{tab:sam-comparison}
\end{table*}

\begin{table}[t]
  \centering
  \setlength{\tabcolsep}{1.25pt}
  \renewcommand{\arraystretch}{0.91}
  \resizebox{\columnwidth}{!}{%
  \begin{tabular}{@{}lcrrrrrr@{}}
    \toprule
    Method & Input & Params & VRAM & GFLOPs & Mean & Median & P95 \\
    \midrule
    \multicolumn{8}{l}{\itshape Official SAM2.1 models with mask-derived box prompts} \\
    SAM2.1 Tiny (mask-derived box) & $1024^2$ & 38.962 & 595.02 & 209.855 & 31.400 & 30.020 & 37.614 \\
    SAM2.1 Small (mask-derived box) & $1024^2$ & 46.060 & 622.10 & 271.644 & 35.286 & 34.167 & 39.915 \\
    SAM2.1 Base+ (mask-derived box) & $1024^2$ & 80.850 & 802.84 & 532.233 & 51.094 & 47.803 & 67.576 \\
    SAM2.1 Large (mask-derived box) & $1024^2$ & 224.447 & 1446.67 & 1622.795 & 119.189 & 96.632 & 236.362 \\
    \addlinespace[1pt]
    \multicolumn{8}{l}{\itshape Retrained SAM variants and adaptations} \\
    EfficientSAM ViT-T & Native & \SparkSecond{10.223} & 461.15 & 208.068 & 34.858 & 33.669 & 42.527 \\
    FastSAM-s & $1024^2$ & 11.790 & 217.76 & 206.375 & 40.252 & 39.240 & 48.126 \\
    HQ-SAM ViT-B & Native & 94.807 & 2810.80 & 981.692 & 91.448 & 90.814 & 97.674 \\
    SAM ViT-B & Native & 94.807 & 2810.80 & 981.692 & 88.707 & 88.313 & 91.649 \\
    SAM-Adapter ViT-B & Native & 93.793 & 3402.49 & 979.005 & 69.314 & 68.966 & 71.192 \\
    SAM-SPL & $256^2$ & 25.510 & \SparkSecond{158.21} & \SparkSecond{29.862} & \SparkSecond{25.497} & \SparkSecond{25.110} & \SparkSecond{27.370} \\
    EfficientTAM & Native & 17.866 & 2271.76 & 228.456 & 92.990 & 86.689 & 132.550 \\
    PicoSAM2 & $96^2$ & \SparkBest{1.263} & \SparkBest{22.43} & \SparkBest{0.647} & \SparkBest{3.277} & \SparkBest{3.200} & \SparkBest{3.653} \\
    RemoteSAM & $896^2$ & 241.590 & 1433.56 & 4025.919 & 580.347 & 574.539 & 635.472 \\
    SAM2-Adapter & Native & 641.272 & 9380.45 & 5982.115 & 236.464 & 235.855 & 242.531 \\
    SAM2-UNeXt & $1024^2$ & 522.120 & 3983.72 & 2567.821 & 117.191 & 117.193 & 117.850 \\
    SAM2-UNet & $352^2$ & 216.530 & 1652.50 & 256.476 & 43.103 & 42.605 & 45.516 \\
    SAMamba & $1024^2$ & 39.542 & 1531.77 & 415.011 & 180.893 & 174.864 & 218.942 \\
    \addlinespace[1pt]
    \multicolumn{8}{l}{\itshape Ours} \\
    \textbf{SPARK-SAM} & $1024^2$ & \textbf{39.688} & \textbf{601.82} & \textbf{215.989} & \textbf{31.451} & \textbf{30.716} & \textbf{37.573} \\
    \bottomrule
  \end{tabular}%
  }
  \caption{Computational efficiency of official SAM2 models and retrained SAM variants/adaptations. Input lists the internal square resolution; Native denotes model-defined dynamic preprocessing. Parameters are in millions, peak VRAM in MiB, computational cost in GFLOPs, and latency in ms. Lower is better.}
  \label{tab:sam-efficiency}
\end{table}

%% file: tables/specialized_comparison.tex
\begin{table}[t]
  \centering
  \setlength{\tabcolsep}{1.25pt}
  \renewcommand{\arraystretch}{0.92}
  \resizebox{\columnwidth}{!}{%
  \begin{tabular}{@{}lrrrrrr@{}}
    \toprule
    \multicolumn{7}{c}{\textbf{NUAA-SIRST}} \\
    Method & IoU$\uparrow$ & nIoU$\uparrow$ & F1$\uparrow$ & Pr$\uparrow$ & Re$\uparrow$ & FA$\downarrow$ \\
    \midrule
    DNANet~\citep{li2023dnanet} & \SparkSecond{76.82} & \SparkSecond{77.12} & \SparkSecond{86.89} & 88.06 & \SparkSecond{85.75} & 70.67 \\
    ACM~\citep{dai2021acm} & 64.11 & 64.00 & 78.13 & 76.61 & 79.71 & 147.83 \\
    SCTransNet~\citep{yuan2024sctransnet} & \SparkBest{78.46} & \SparkBest{80.04} & \SparkBest{87.93} & \SparkBest{94.83} & 81.96 & \SparkBest{27.14} \\
    RPCANet++~\citep{wu2025rpcanetpp} & 63.48 & 62.96 & 77.66 & 78.00 & 77.33 & 132.46 \\
    L2SKNet~\citep{wu2024l2sknet} & 55.27 & 53.47 & 71.19 & 82.41 & 62.66 & 81.25 \\
    BGM~\citep{liu2025forgetting} & 71.46 & 75.02 & 83.35 & 88.84 & 78.51 & 59.92 \\
    NS-FPN~\citep{yuan2026seeing} & 76.02 & 76.40 & 86.38 & \SparkSecond{90.54} & 82.58 & 52.40 \\
    ISNet~\citep{zhang2022isnet} & 70.73 & 68.57 & 82.86 & 90.51 & 76.40 & \SparkSecond{48.65} \\
    \addlinespace[1pt]
    \textbf{SPARK-SAM} & \textbf{75.78} & \textbf{76.93} & \textbf{86.22} & \textbf{84.18} & \SparkBest{88.37} & \textbf{100.88} \\
    \midrule
    \multicolumn{7}{c}{\textbf{NUDT-SIRST}} \\
    Method & IoU$\uparrow$ & nIoU$\uparrow$ & F1$\uparrow$ & Pr$\uparrow$ & Re$\uparrow$ & FA$\downarrow$ \\
    \midrule
    DNANet~\citep{li2023dnanet} & 84.89 & 85.69 & 91.83 & 90.40 & 93.31 & 66.33 \\
    ACM~\citep{dai2021acm} & 40.40 & 36.39 & 57.55 & 58.14 & 56.97 & 274.49 \\
    SCTransNet~\citep{yuan2024sctransnet} & \SparkBest{94.42} & \SparkBest{94.76} & \SparkBest{97.13} & \SparkBest{98.58} & \SparkSecond{95.72} & \SparkBest{9.21} \\
    RPCANet++~\citep{wu2025rpcanetpp} & \SparkSecond{93.74} & 94.05 & \SparkSecond{96.77} & 97.22 & \SparkBest{96.32} & 18.43 \\
    L2SKNet~\citep{wu2024l2sknet} & 91.62 & 92.86 & 95.63 & 96.48 & 94.79 & 23.15 \\
    BGM~\citep{liu2025forgetting} & 93.39 & \SparkSecond{94.12} & 96.58 & \SparkSecond{97.92} & 95.28 & \SparkSecond{13.53} \\
    NS-FPN~\citep{yuan2026seeing} & 77.63 & 80.74 & 87.40 & 88.93 & 85.93 & 71.63 \\
    ISNet~\citep{zhang2022isnet} & 58.52 & 63.68 & 73.83 & 71.56 & 76.25 & 202.80 \\
    \addlinespace[1pt]
    \textbf{SPARK-SAM} & \textbf{86.49} & \textbf{87.75} & \textbf{92.75} & \textbf{93.63} & \textbf{91.90} & \textbf{41.86} \\
    \midrule
    \multicolumn{7}{c}{\textbf{IRSTD-1K}} \\
    Method & IoU$\uparrow$ & nIoU$\uparrow$ & F1$\uparrow$ & Pr$\uparrow$ & Re$\uparrow$ & FA$\downarrow$ \\
    \midrule
    DNANet~\citep{li2023dnanet} & 66.05 & 58.88 & 79.56 & 79.88 & 79.24 & 50.16 \\
    ACM~\citep{dai2021acm} & 50.29 & 33.29 & 66.92 & 69.41 & 64.61 & 71.56 \\
    SCTransNet~\citep{yuan2024sctransnet} & \SparkBest{69.89} & \SparkBest{65.75} & \SparkBest{82.27} & \SparkSecond{83.24} & \SparkBest{81.33} & \SparkSecond{41.16} \\
    RPCANet++~\citep{wu2025rpcanetpp} & 61.74 & 53.82 & 76.34 & 75.51 & 77.19 & 62.90 \\
    L2SKNet~\citep{wu2024l2sknet} & 67.45 & 61.84 & 80.56 & 81.68 & 79.47 & 44.78 \\
    BGM~\citep{liu2025forgetting} & 67.05 & \SparkSecond{65.48} & 80.27 & 81.05 & 79.51 & 46.71 \\
    NS-FPN~\citep{yuan2026seeing} & 65.07 & 59.56 & 78.84 & \SparkBest{83.68} & 74.53 & \SparkBest{36.53} \\
    ISNet~\citep{zhang2022isnet} & 65.11 & 58.94 & 78.87 & 78.20 & 79.55 & 55.73 \\
    \addlinespace[1pt]
    \textbf{SPARK-SAM} & \SparkSecond{68.34} & \textbf{64.47} & \SparkSecond{81.19} & \textbf{81.53} & \SparkSecond{80.86} & \textbf{46.02} \\
    \bottomrule
  \end{tabular}%
  }
  \caption{Comparison with specialized IRSTD networks. IoU, nIoU, F1, precision (Pr), and recall (Re) are percentages; FA is in px/MP. Cyan and orange denote the best and second-best results.}
  \label{tab:specialized-comparison}
\end{table}

%% file: tables/mechanism_analysis.tex
\begin{table}[t]
  \centering
  \begingroup
  \setlength{\tabcolsep}{1.7pt}
  \renewcommand{\arraystretch}{0.92}
  \scriptsize

  \begin{tabular*}{\columnwidth}{@{\extracolsep{\fill}}lrrrrrr@{}}
    \toprule
    \multicolumn{7}{l}{\itshape (a) Prompt interventions: change from learned prompts} \\
    & \multicolumn{2}{c}{NUAA} & \multicolumn{2}{c}{NUDT} & \multicolumn{2}{c}{IRSTD} \\
    \cmidrule(lr){2-3}\cmidrule(lr){4-5}\cmidrule(lr){6-7}
    Intervention & $\Delta$IoU & $\Delta$nIoU & $\Delta$IoU & $\Delta$nIoU & $\Delta$IoU & $\Delta$nIoU \\
    \midrule
    No prompt state        & $-74.70$ & $-28.27$ & $-41.74$ & $-8.84$ & $-47.09$ & $-18.68$ \\
    Random spatial branch  & $-74.84$ & $-33.63$ & $-14.26$ & $-5.45$ & $-18.65$ & $-0.65$ \\
    Shuffled spatial branch& $-74.95$ & $-29.41$ & $-5.39$  & $-5.15$ & $-0.11$  & $-0.14$ \\
    Box-only shuffle   & $-0.49$  & $-0.11$  & $-0.45$  & $-0.18$ & $-0.20$  & $-0.20$ \\
    Mask-derived box & $+1.57$  & $+0.92$  & $+2.48$  & $+0.30$ & $+1.77$  & $+2.22$ \\
    \bottomrule
  \end{tabular*}

  \vspace{3pt}
  \begin{tabular*}{\columnwidth}{@{\extracolsep{\fill}}lrrrrr@{}}
    \toprule
    \multicolumn{6}{l}{\itshape (b) Contributions of response guidance and prompt refinement} \\
    & Full & \multicolumn{2}{c}{w/o guidance} & \multicolumn{2}{c}{w/o refinement} \\
    \cmidrule(lr){3-4}\cmidrule(lr){5-6}
    Dataset & IoU & IoU & $\Delta$IoU & IoU & $\Delta$IoU \\
    \midrule
    NUAA  & $75.78$ & $61.20$ & $-14.58$ & $75.28$ & $-0.50$ \\
    NUDT  & $86.49$ & $62.62$ & $-23.87$ & $85.44$ & $-1.05$ \\
    IRSTD & $68.34$ & $55.99$ & $-12.35$ & $66.82$ & $-1.52$ \\
    \bottomrule
  \end{tabular*}
  \endgroup
  \caption{Mechanism analysis. All changes use variant-minus-full-model percentage points. (a) applies frozen-weight interventions to the joint self-prompt state: spatial rows preserve $T_{\mathrm{res}}$ and box context, box rows preserve $T_{\mathrm{res}}$ and the spatial branch, and mask-derived box replacement substitutes, after adaptation, a box computed from the test reference mask. (b) reports matched training ablations that separately remove the complete response-guidance pipeline and high-resolution prompt refinement.}
  \label{tab:mechanism-analysis}
\end{table}